\documentclass[11pt]{article}

\usepackage[margin=1in]{geometry}
\usepackage{graphicx}
\usepackage{amsmath,amssymb}
\usepackage{booktabs}
\usepackage{multirow}
\usepackage{subcaption}
\usepackage[round,authoryear]{natbib}
\usepackage{authblk}
\usepackage{xcolor}
\usepackage[hidelinks]{hyperref}

\title{Can the Waymo Open Motion Dataset Support Realistic Behavioral Modeling? A Validation Study with Naturalistic Trajectories}

\author[1]{Yanlin Zhang}
\author[1]{Sungyong Chung}
\author[2]{Nachuan Li}
\author[2]{Dana Monzer}
\author[2]{Hani S. Mahmassani\thanks{Deceased.}}
\author[3]{Samer H. Hamdar}
\author[1]{Alireza Talebpour\thanks{Corresponding author, \href{mailto:ataleb@illinois.edu}{ataleb@illinois.edu}}}

\affil[1]{Department of Civil and Environmental Engineering, University of Illinois at Urbana-Champaign}
\affil[2]{Northwestern University Transportation Center}
\affil[3]{Department of Civil and Environmental Engineering, George Washington University}

\date{}

\begin{document}
\maketitle

\begin{abstract}
\noindent The Waymo Open Motion Dataset (WOMD) has become a popular resource for data-driven modeling of autonomous vehicles (AVs) behavior. However, its validity for behavioral analysis remains uncertain due to proprietary post-processing, the absence of error quantification, and the segmentation of trajectories into 20-second clips. This study examines whether WOMD accurately captures the dynamics and interactions observed in real-world AV operations. To achieve this objective, this paper introduces a robust, error-aware validation methodology for assessing whether a public vehicle-trajectory dataset is suitable for driving behavior and traffic-flow research using an independently collected ground-truth reference dataset with a characterized error distribution. Accordingly, leveraging an independently collected naturalistic dataset from Level 4 AV operations in Phoenix, Arizona (PHX), we perform comparative analyses across three representative urban driving scenarios: discharging at signalized intersections, car-following, and lane-changing behaviors. For the discharging analysis, headways are manually extracted from aerial video to ensure negligible measurement error. For the car-following and lane-changing cases, we apply the Simulation–Extrapolation (SIMEX) method to account for empirically estimated error in the PHX data and use Dynamic Time Warping (DTW) distances to quantify behavioral differences. Results across all scenarios consistently show that behavior in PHX falls outside the behavioral envelope of WOMD. Notably, WOMD underrepresents short headways and abrupt decelerations. These findings suggest that behavioral models calibrated solely on WOMD may systematically underestimate the variability, risk, and complexity of naturalistic driving. Caution is therefore warranted when using WOMD for behavior modeling without proper validation against independently collected data.

\end{abstract}


\section{Introduction}
Autonomous vehicles (AVs), once considered futuristic, are now operating commercially in many cities. Over the past 40 years, numerous studies have proposed optimized AV car-following \citep{talebpour2016influence,mahdinia2021integration} and lane-changing \citep{an2023optimized,an2023vehicle} models aimed at improving traffic flow efficiency. The majority of these studies focus on lower levels of autonomy, including Adaptive Cruise Control (ACC) and automated lane-changing. Moreover, most of these studies rely on simplifying assumptions about the AV software stack. For example, detailed vehicle dynamics models are rarely incorporated when analyzing string stability, and perception challenges are often reduced to structured noise models. Such simplifications are in stark contrast to the realities of AV deployment, as the industry moves towards the deployment of AVs at high levels of automation.

In practice, AV development has been shaped more by safety, reliability, regulatory compliance, and user comfort than by theoretical optimality (often at the cost of efficiency). Traditional robotic-based architectures used in commercial AVs leave little room for any simplifications, as they must ensure robust operation under diverse and uncertain conditions, even if that requires suboptimal decisions due to computational constraints and safety requirements. Compounding this divergence, recent foundation model-based approaches largely prioritize reproducing observed human driving behavior over optimizing for traffic efficiency, abandoning the notion of optimality in favor of compatibility with human driver expectations. Examples include large Transformer-based motion forecasting models such as Wayformer \citep{nayakanti2022wayformer}, vectorized end-to-end driving stacks such as VAD \citep{jiang2023vad,chen2024vadv2}, and emerging surveys of foundation models for autonomous driving perception and planning \citep{sathyam2025foundation,gao2024survey}. These systems are trained on vast datasets and optimize multi-task objectives, but they inherit any measurement and post-processing biases present in the underlying datasets.

Accordingly, several recent studies have deviated from focusing on optimality and utilized real-world data from AV operations to model AV behavior \citep{hu2022processing}, assess its impacts on traffic flow dynamics \citep{hu2023autonomous}, and characterize the impacts of AVs on human behavior in their vicinity \citep{zhang2023influence,rahmati2019influence}. The majority of these studies have utilized one of three available datasets from real-world AV operations at high levels of automation: Waymo Open Motion Dataset (WOMD) \citep{ettinger2021large,chen2024womd}, Lyft level-5 Open Dataset \citep{houston2021one}, and Argoverse Dataset \citep{chang2019argoverse}. Considering that the Lyft Level-5 Open Dataset and Argoverse Dataset are no longer being updated, WOMD has become a dominant dataset for such analyses. Although a wide range of topics have been covered in studies that utilized WOMD, they all have one common underlying assumption: WOMD is accurate and suitable for behavioral and traffic flow research. Unfortunately, this dataset offers no avenue for independent accuracy validation. 

% Focusing on the WOMD, there exist two sources of error in this dataset: (1) Ego vehicle poses are derived from IMU/GNSS and are therefore an order of magnitude more precise than the LiDAR, camera, and radar fusion used for surrounding objects. Researchers cannot quantify the error structure because the raw sensor data and the technical details for data processing are withheld, and (2) all published trajectories are the outcome of a proprietary off-board tracking pipeline that implicitly applies temporal smoothing across several past and future frames for regressing the current bounding box. While this filtering is advantageous for motion planning benchmarks, as it delivers "clean" inputs with reduced measurement noise, it can potentially suppress exactly the discontinuous elements that characterize real human driving, for example, stop–and-go oscillations in queuing, start-up reaction at signalized intersections, and the lateral jerk during lane-changing behaviors. 

Despite its breadth and popularity, the Waymo Open Motion Dataset (WOMD) has several structural limitations when used for microscopic behavioral analysis. First, WOMD is constructed from 20~s segments for interactions and then further cropped to 9~s scenarios for the Interaction Prediction Challenge \footnote{The details for Waymo Open Dataset Interaction Prediction Challenge can be found at \url{https://waymo.com/open/challenges/2025/interaction-prediction/}}, rather than preserving long, continuous runs of naturalistic driving. This design is ideal for motion-forecasting challenges, but it truncates car-following and lane-changing histories and removes information about how drivers adapt over multiple signal cycles, merges, and shockwaves. As a result, key state variables such as long-range headway history, multi-episode stopping and re-acceleration, or sequence of lane changes cannot be reconstructed, which limits the calibration and validation of dynamic behavioral models that inherently depend on a longer temporal context.

Second, the released WOMD scenarios are effectively outdated. Since the initial public release in 2021, the dataset has contained the same 103{,}354 20~s scenarios; subsequent versions only add new modalities, including enhanced maps, LiDAR point clouds, camera embeddings, and traffic signals for these existing clips, without introducing new driving episodes. Consequently, the trajectories predominantly reflect autonomous-driving policies deployed before 2021, whereas the commercial operations of the major Level~4 ride-hailing provider in Phoenix have undergone substantial software and policy updates since then. In other words, even if WOMD trajectories were perfectly measured, they still represent an outdated slice of autonomous-vehicle behavior.

Third, all trajectories in WOMD are the output of an offboard perception pipeline that explicitly aggregates information over a point-cloud sequence, including both past and future frames, to estimate the current bounding box of an object and generate temporally smooth 3D tracks \citep{qi2021offboard,ettinger2021large}. This offline smoothing greatly reduces perception noise and tracking fragmentation but also suppresses exactly the kinds of discontinuities that are critical to behavioral and traffic-flow modeling, such as start-up delays at signals, stop-and-go oscillations in queues, abrupt hazard-induced braking, or lateral oscillation during lane changes. Any model calibrated directly on these smoothed trajectories will tend to understate variability in acceleration, gap acceptance, and reaction times. Moreover, recent work has shown that even non-trajectory portions of WOMD, such as traffic signal states, exhibit substantial missing and inaccurate entries with 71.7\% of signal states unknown or missing in the raw data and requiring substantial post hoc imputation and correction \citep{yan2026improving}.

Accordingly, this study aims to answer one fundamental question regarding WOMD: Can this dataset support behavioral and traffic flow research? To answer this question, the observed behavior in the WOMD is compared to an independently collected naturalistic dataset from Level 4 AV operations in Phoenix, AZ (PHX). The analysis focuses on three representative driving scenarios: intersection discharging, car-following, and lane-changing. For the intersection discharging case, headways are manually extracted from high-resolution aerial videos in the PHX dataset with negligible error. These manually extracted headways are then compared against those extracted from WOMD trajectories at intersections, which rely on vehicle pose estimates and map alignment. For the car-following and lane-changing scenarios, we analyze trajectories from both datasets with error-aware statistical tools to assess their alignment. In particular, we employ the Simulation–Extrapolation (SIMEX) procedure~\citep{cook1994simulation} to remove the effect of empirically estimated measurement noise in the PHX dataset. This correction allows for a fair comparison of behavior patterns such as decelerating-to-stop responses and lane-change interactions. By quantifying cross-dataset differences and comparing them against the internal variability of WOMD, we assess whether the naturalistic behaviors observed in the PHX dataset fall within the behaviors captured in WOMD. 

It is important to note that this is not a perfectly controlled experiment. The PHX trajectories are collected in 2023 from a specific corridor in Phoenix under normal driving scenarios, whereas WOMD aggregates self-driving operations from multiple U.S. cities before 2021. Moreover, commercial AV behavior evolves rapidly. Even if WOMD trajectories were free of measurement and post-processing error, a nontrivial portion of the discrepancy we observe may reflect (i) changes in the underlying driving policy over time, (ii) differences in roadway geometry and traffic composition across cities, and (iii) the use of different sensing and smoothing pipelines, for example, aerial videography versus on-board, proprietary perception and tracking. Accordingly, our goal is not to attribute discrepancies solely to a single source, but rather to quantify whether recent, naturalistic Level~4 operation in Phoenix lies within the behavioral envelope spanned by WOMD, and in doing so, to critically assess the validity of using WOMD directly for driving behavior studies.

\textcolor{black}{It is also important to note that several studies have examined the accuracy and internal consistency of vehicle trajectory datasets, particularly in the context of the Next Generation Simulation (NGSIM) program \citep{punzo2011assessment, montanino2013making,montanino2015trajectory}. Among the most influential contributions, \citet{punzo2011assessment} proposed a comprehensive framework for assessing trajectory data quality based on kinematic consistency diagnostics, including jerk distributions, acceleration smoothness, and spectral properties of vehicle motion. In that work, a Kalman-filtered reconstruction of the original trajectories was used as a reference to quantify inconsistency in the raw NGSIM data. While this approach established important benchmarks for identifying measurement noise and post-processing artifacts, it relies on internally reconstructed trajectories derived from the same data source rather than an external ground-truth reference. As a result, the assessment primarily evaluates internal consistency and physical plausibility, rather than absolute validity with respect to independently observed vehicle motion. Subsequent studies adopting similar principles have likewise focused on smoothing-based corrections and internal diagnostics \citep{hu2022processing,hu2023IEEE}, reflecting the limited availability of independent, high-resolution ground truth in large-scale trajectory datasets.} \textcolor{black}{In contrast, the present study advances trajectory dataset validation by leveraging an independently collected naturalistic dataset with empirically characterized measurement error, enabling an external and error-aware assessment of a public autonomous-vehicle trajectory dataset.}

\textcolor{black}{Considering the abovementioned discussions, this study makes two contributions: First, we introduce a robust methodology to assess the validity of a vehicle trajectory dataset. Second, leveraging a self-collected naturalistic robotaxi dataset (PHX) with empirically characterized error, to our knowledge, we provide the first independent validation of WOMD using third-party data (serving as the ground truth) rather than relying on WOMD’s own post-processed outputs as a reference.}

The remainder of this paper is organized as follows. Section 2 describes the two datasets and preprocessing steps, including error estimation for the naturalistic PHX dataset. Section 3 investigates discharge headways at signalized intersections. Section 4 presents the analysis of car-following behavior through a probabilistic state transition framework and a SIMEX-corrected dynamic time warping (DTW) comparison. Section 5 evaluates lane-changing behavior using a multivariate DTW distance and permutation testing. Finally, Section 6 concludes with insights on the implications of our findings for behavioral modeling using AV datasets such as WOMD.

\section{Data Description} \label{sec:data}

The empirical foundation of this study rests on two complementary sources of vehicle‐trajectory data.  % The first is an independently collected naturalistic dataset from Phoenix, Arizona (PHX), captured via high-resolution aerial videography and processed with a multi-object tracking pipeline to recover submeter-level trajectories. The second is the WOMD, a large-scale benchmark generated from an industry-grade perception stack onboard autonomous vehicles. Juxtaposing these datasets enables a rigorous assessment of whether motion patterns inferred from WOMD faithfully replicate the kinematic and interaction dynamics observed in a real-world setting, as in the PHX dataset.  
The next two subsections describe the collection protocols, preprocessing steps, and key descriptive statistics for each dataset, beginning with the Naturalistic PHX Dataset.

\subsection{Naturalistic Phoenix Dataset}
\label{sec:phx_data}
To achieve the objective mentioned above, this section introduces the data collection and trajectory extraction processes used in this study. Additionally, it includes an analysis of the accuracy of the PHX dataset.

\subsubsection{Data collection}
The first step is to perform a comprehensive data collection. The goal was to capture uninterrupted AV driving in real-world conditions. The two main locations selected for data collection are Phoenix and Chandler, AZ. The selected locations, specifically downtown Phoenix, allow for the existence of a series of challenging real-world scenarios that the AVs have to navigate. Data collection was performed in late January 2023, in clear, sunny weather conditions, between 11 am and 5 pm. Sunny weather conditions usually mean increased shadows in the collected data, which makes data post-processing steps more challenging. The data collection method used in this work is based on the Moving Aerial Videography \citep{TGSIM,TGSIMreport}, which includes a helicopter mounted with an 8K resolution camera that collects birds-eye view footage at 500 feet altitude by following the vehicles of interest throughout the study duration. High-quality, information-rich interactions are clipped into smaller segments for later processing and analysis. The behavior of AVs is captured during straight road segment movements, left/right turns, at intersections, and within several other driving scenarios. The ego vehicles targeted in this data collection are driverless robotaxis offered as a public ride service by a single major provider of automated mobility in Phoenix and Chandler.\footnote{According to public statements and SAE J3016, these services operate without an on-board safety driver within a geo-fenced operational design domain that corresponds to Level~4 automated driving capability.} Throughout the study runs, these vehicles operated without any human intervention inside the vehicle cabin. They performed everyday driving tasks such as car-following, lane-changing, navigating signalized intersections, and yielding to pedestrians and other road users. All surrounding traffic in the PHX dataset consists of conventional human-driven vehicles sharing the roadway with the ego robotaxis.Several runs of data collection are performed, where a complete helicopter run consists of approximately 12 miles of road navigated on the ground.

\subsubsection{Trajectory extraction}
Vehicle trajectory information is later extracted from the recorded videos using the methodology explained in \citep{TGSIM,TGSIMreport}. The method begins by extracting high-resolution raw images from the video recordings, which are recorded at 30 frames per second. Next, a few representative frames are selected to create a reference image of the study area, which is necessary to transform the coordinates of each image into a global frame of reference. The next steps include object detection and tracking, where vehicles in each frame are detected using a RetinaNet-based detector that was retrained based on the data collected in this study \citep{hosseini2022unsupervised}. Object tracking includes assigning a unique ID to each vehicle using data association and track maintenance \citep{hosseini2022unsupervised}. 
Subsequently, image stabilization is performed to ultimately transform each frame to the global coordinate system.% of the reference image using a combination of algorithms (i.e., Scale-Invariant Feature Transform (SIFT) \citep{lindeberg2012scale}, Fast Library for Approximate Nearest Neighbors (FLANN) \citep{muja2009flann}, and Random Sample Consensus (RANSAC) \citep{fischler1981random}). The algorithms are used to first detect the key features in both the reference image and input images; next, features between those images are matched before finally estimating the transformation between them using a homography matrix. 
Once transformed into a global frame, trajectories are extracted.

\begin{figure}[!htb]
    \centering
    \includegraphics[width=\textwidth]{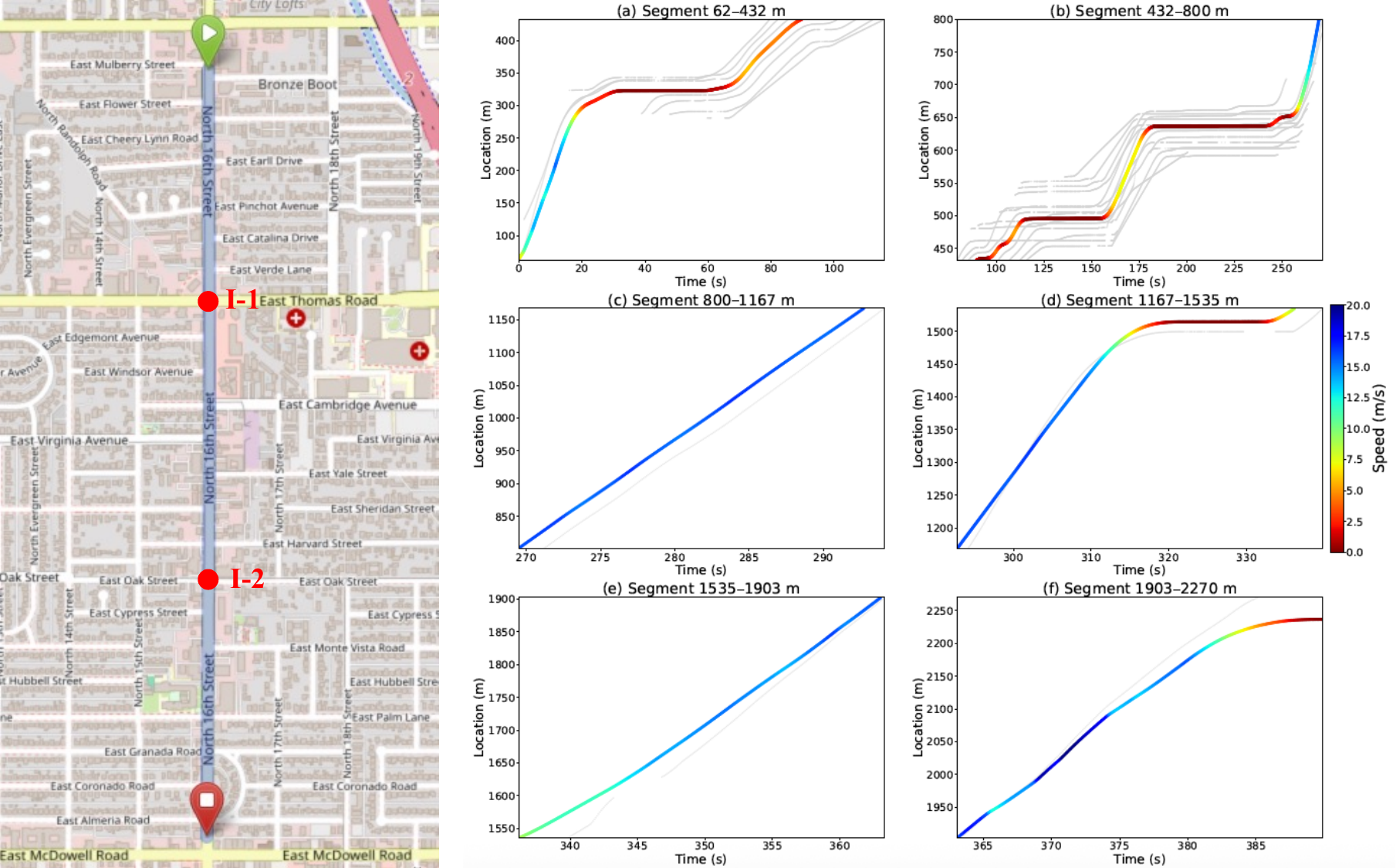} 
    \caption{Reconstructed vehicle trajectories along a 2.2-km segment of southbound N 16th St. in Phoenix, AZ, during the afternoon peak hour. The autonomous vehicle (AV) trajectory is color-coded by speed, while all other vehicles on the two main through lanes are shown in light gray. The corridor includes two signalized intersections: I-1 (N 16th St. \& E Thomas Rd., at $x = 650$~m) and I-2 (N 16th St. \& E Oak St., at $x = 1500$~m). (Map data © OpenStreetMap. License: \url{https://www.openstreetmap.org/copyright})}
    \label{fig:trajectory_reconstruction}
\end{figure}

Figure \ref{fig:trajectory_reconstruction} illustrates the reconstructed vehicle trajectories traveling southbound on N 16th St. in Phoenix, AZ, during the afternoon peak hour. The AV is highlighted with a speed color-coding, while the trajectories of other vehicles on the two main lanes are shown in light grey. Vehicles on the left-turn lanes at intersections are not included. This corridor features two signalized intersections. Due to heavy traffic, vehicles experienced multiple stops when approaching the first traffic signal at N 16th St. \& Thomas Rd. The total length of this section is approximately 2,200 meters, starting from E Osborn Rd., where the AV turns onto N 16th St., and ending at E Almeria Rd., where the AV makes a right turn.

\subsubsection{Error Estimation}
\label{sec:phx_error}
To evaluate the accuracy of the extracted trajectories and assess the suitability of this dataset for analyzing microscopic driving behavior dynamics, we estimated the distribution of travel distance error rather than instantaneous positional error. This distinction is important: due to the object tracking pipeline applied to aerial video, positional estimates are smoothed and spatially correlated, whereas distance traveled, computed as displacement over a time interval, is more relevant for dynamic behavior analysis.

To generate ground truth travel distances, we manually identified 40 distinct control points along a 2.2-km southbound corridor (Figure \ref{fig:trajectory_reconstruction}). The $(x,y)$ coordinates in the PHX dataset are defined in a global frame constructed by transforming the aerial frames into a high-resolution reference image, which is geometrically matched to a geo-referenced satellite image of the study area. This matching aligns the axes of the reference image with UTM Easting and Northing \citep{snyder1987map}, so that motion in $x$ and $y$ closely matches the change in UTM coordinates. The downtown Phoenix experiment area is laid out as an almost perfectly rectilinear grid with very small curvature within the data collection area, so roadway centerlines are approximately parallel or orthogonal to these axes. For each corridor, we therefore define the longitudinal direction along the predominant axis of travel and the lateral direction along the orthogonal axis. 

For each vehicle trajectory, we reviewed raw video frames to determine the exact time at which the vehicle center passed each control point. Frame numbers were converted into timestamps based on the video's frame rate at 30 fps, and all valid time-point pairs were identified. The resulting ground truth is therefore accurate up to the limits imposed by the video frame rate and pixel resolution in the 8K imagery on the order of 0.0333 seconds and 0.3 meters. This construction makes the PHX dataset a trajectory set with empirically characterized error: the residuals between aerially extracted trajectories and the manually labeled reference distances are precisely what our error model captures. To obtain a representative distribution of travel distances, we employed stratified random sampling without replacement across four trajectory duration quartiles: intervals below 16.9 seconds, between 16.9–44.8 seconds, 44.8–80.6 seconds, and above 80.6 seconds. This stratification reflects the expectation that tracking error may increase with the length and duration of the interval; it also prevents very short hops from dominating the error sample. A total of 60 pairs were selected. For each pair, the measured longitudinal and lateral distances were compared to ground truth to obtain the 2D distance-traveled error vector. 

\begin{figure}[h]
    \centering
    \includegraphics[width=0.8\textwidth]{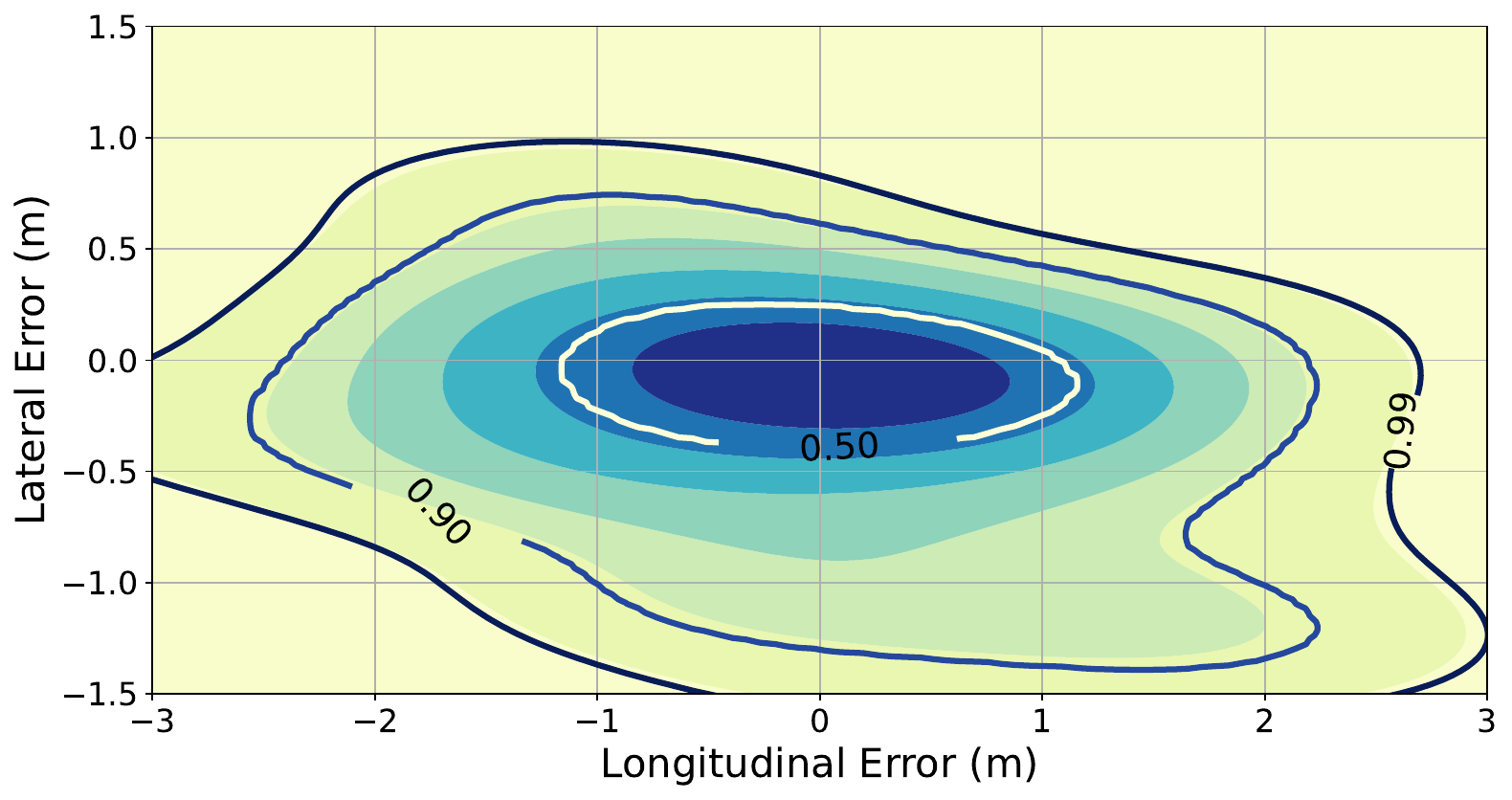} 
    \caption{Travel distance error distribution. The contour lines show the proportion of the data within the range.}
    \label{fig:distance_error}
\end{figure}

The resulting distribution of travel distance error was modeled using a bivariate normal distribution. As shown in figure \ref{fig:distance_error}, the final error distribution, based on 60 travel distance samples, yielded the following parameters: mean longitudinal error $\mu_x = 0.276$ m, mean lateral error $\mu_y = 0.006$ m, standard deviations $\sigma_x = 1.075$ m and $\sigma_y = 0.530$ m, and a moderate negative correlation $\rho_{xy} = -0.291$. Mardia's multivariate skewness and kurtosis tests returned $p = 0.104$ and $p = 0.580$, respectively, indicating that the null hypothesis of joint normality cannot be rejected. From this travel distance error model, we derived the distribution of spacing error between two vehicles by taking the difference of two independent errors. Assuming independence, the spacing error follows a distribution $g \sim \mathcal{N}(0,\;2\sigma_x^2) = \mathcal{N}(0,\;2.309\;\text{m}^2)$. 

% However, this spacing error is relevant only over the typical time interval used for estimation (median $\Delta t \approx 4.5$ seconds). To model the instantaneous spacing error at the trajectory's native temporal resolution (0.1 s), we assume a random-walk accumulation process, commonly adopted in aerial vehicle tracking literature \citep{apeltauer2015automatic, shi2025consistent}, such that $\text{Var}[e_x(\Delta t)] = q\Delta t$. Based on the observed error variance and average segment duration, we estimated $q \approx 0.25\;\text{m}^2/\text{s}$. This implies an instantaneous position error variance of $0.025\;\text{m}^2$ at 0.1 s, and a spacing error variance of $0.051\;\text{m}^2$, or standard deviation $\approx 0.226$ m—well within the size of a vehicle.

We further derived the speed error distribution by dividing the travel distance error by the elapsed time for each trajectory. The estimation results in a space-mean speed error distribution with mean $\mu_v = [0.0163,\;0.0004]$ m/s, standard deviations $\sigma_{v_x} = 0.0636$ m/s and $\sigma_{v_y} = 0.0314$ m/s, and correlation $\rho_{v_xv_y} = -0.183$. Mardia's skewness and kurtosis test on this distribution yield $p = 0.34$ and $p = 0.58$, confirming that the speed error is well approximated by a 2D Gaussian. Finally, the relative speed error between a lead vehicle and its follower was derived as the difference of two independent speed errors. The resulting distribution is zero-mean, with standard deviations $\sigma_{\Delta v_x} = 0.0899$ m/s and $\sigma_{\Delta v_y} = 0.0444$ m/s, and the same correlation $\rho = -0.183$. This formulation ensures consistent propagation of uncertainty from trajectory error to car-following dynamics in a behaviorally interpretable manner.

\subsection{Waymo Open Motion Dataset}
The WOMD captures a variety of scenarios, including lane-changing events involving Waymo vehicles surrounded by HVs, as well as those involving only HVs. With its extended lidar detection range, the dataset can also capture lane-changing events that occur without any interaction with Waymo vehicles. Since the data is collected by sensors mounted on Waymo vehicles, all scenarios include Waymo's trajectory as well as the trajectories of surrounding objects detected within the sensor's range at each time step. According to the WOMD homepage, the data is collected from various locations, including San Francisco, Mountain View, Los Angeles, Detroit, Seattle, and Phoenix. For this study, we specifically used the March 2023 dataset.

The motion dataset offers 20-second segments sampled at 10 Hz, representing a variety of scenarios along with their corresponding map data. The dataset includes details such as a unique scenario ID, unique tracking IDs for each object, object types (i.e., vehicles, pedestrians, and cyclists), whether a vehicle is Waymo or not, and attributes like object position (x, y, z), dimensions (length, width, height), heading, and velocity. For each scenario, the dataset also offers detailed map information, including lane centers, lane boundaries, and road boundaries, represented as 3D polylines or polygons. Further details about the motion dataset are available at (\url{https://waymo.com/open/data/motion/}).

A key distinction is that the PHX trajectories are derived from a stabilized, top-down aerial viewpoint, whereas WOMD trajectories are produced by an ego-centric perception and tracking stack mounted on the self-driving vehicle and by offboard processing. As a result, PHX errors are dominated by object-tracking uncertainty in orthorectified imagery, whereas WOMD errors are driven by on-board sensor noise, occlusions, and proprietary temporal smoothing.

In summary, WOMD provides a broad but temporally and spatially heterogeneous snapshot of AV operations, while the PHX dataset offers a smaller but more homogeneous sample of everyday robotaxi driving in a specific urban corridor in early 2023. The combination of these two views allows us to assess whether typical maneuvers observed in up-to-date commercial service are well represented in the public dataset.

\section{Intersection Discharge Headway Analysis} \label{sec:headway}

%Understanding the differences in discharge headways at an intersection is important because it allows more understanding of how the existence of AVs can affect road capacity. By leveraging assumptions about the driving behaviors of AVs, numerous studies have demonstrated that AVs can improve intersection capacity through theoretical derivation and simulation \citep{yu2023impact, park2021impact, song2023intersection}. However, it is unknown whether this conclusion is consistent with what was observed in the naturalistic driving environment. WOMD, featuring numerous intersections where a large sample of starting-up trajectories of AVs and HVs are collected, seems to be an appealing source to verify these conclusions. Unfortunately, due to its data-smoothing technique,  WOMD may fail to capture the discontinuity in the driving behavior of vehicles near intersections. Therefore, it is unclear whether WOMD enables unbiased analysis of vehicles' discharge behavior near the intersection for through movement. In this section, we compare the distributions of discharge headways extracted in WOMD and PHX and observe whether significant differences exist between them.

Understanding the vehicles' discharge headways at an intersection is crucial because it enables understanding of the road capacity. WOMD features numerous intersections where a large sample of starting-up trajectories of AVs and HVs are collected. Accordingly, WOMD is an appealing source to examine the effects of AVs on road capacity, which is a conclusion drawn by numerous existing studies through theoretical derivation and simulation \citep{yu2023impact, park2021impact, song2023intersection}. Unfortunately, due to its data-smoothing technique, WOMD may fail to capture the discontinuity in the driving behavior of vehicles near intersections. Therefore, it is unclear whether WOMD enables an unbiased analysis of vehicles' discharging behavior when the signal turns green.

In this section, we assess the accuracy of through-movement discharge headways in WOMD by comparing the overall distribution to the PHX dataset and the headway compression phenomenon to an existing study \citep{lin2005headway}.

\subsection{Discharge Headway Extraction}

To make sure that the collected headways are representative of the discharging process, we ensure that the collected time when vehicles start from intersections is from those that are already waiting at the intersection. This is to prevent biasing our analysis with free-flow headways. In addition, lanes that allow left-turns and the portion of queues behind right-turning vehicles are excluded, since turning vehicles tend to significantly lower their speeds and adopt longer headways \citep{manual2000highway}.
We also note that due to the lack of signal information, we extract discharge headways starting from the second vehicle in the queue. Given the distinct data collection methods, headway extraction methods are customized for the PHX and WOMD datasets. 

\subsubsection{PHX Dataset}
% what is different for the methods? video vs smoothed trajectory
In the PHX dataset, we inspect each frame of the raw videos at a resolution of 10 frames per second and obtain the start time of each vehicle from the intersection. This method ensures limited measurement error compared to the ground truth, which means that the collected headways can serve as a relatively reliable reference, despite the relatively low sample size.

\begin{figure}[!htb]
    \centering
    \includegraphics[width=0.7\textwidth]{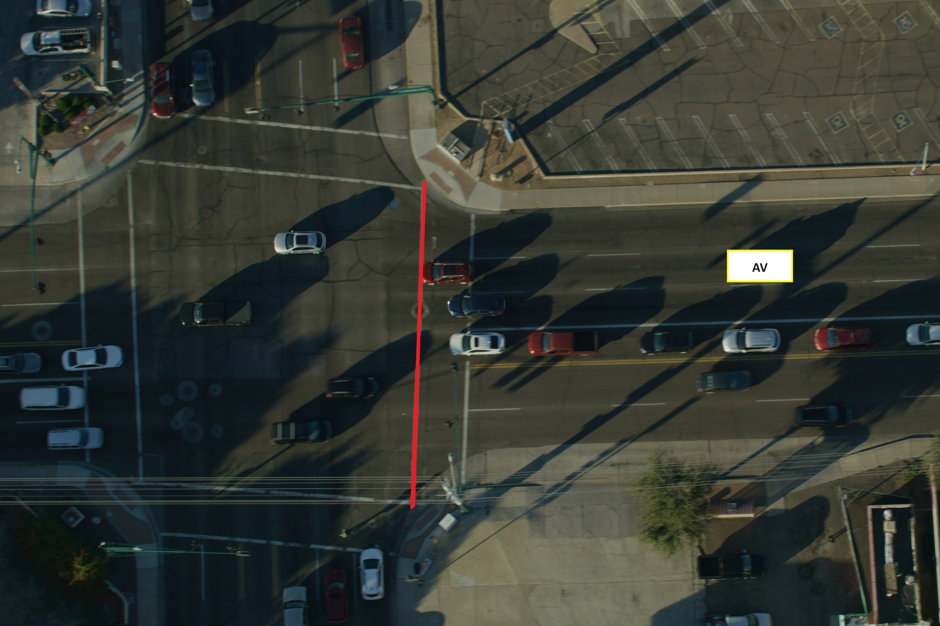} % example-image is a placeholder; replace with your image file name
    \caption{PHX Dataset Intersection Geometry}
    \label{intersection}
\end{figure}

In all, we identified 30 through-movement lanes with at least two discharging vehicles, and 10 of these lanes contain an AV. Figure \ref{intersection} shows the geometry of an intersection in the PHX dataset. For each vehicle, we obtain the frame when its front bumper is closest to the red line in Figure \ref{intersection}. The stop line is not used as the reference line, because some vehicles' front bumpers may be downstream of it while waiting. After obtaining the start times, we calculate the discharge headway by subtracting those of the adjacent vehicles. This procedure yields headway measurements whose uncertainty is effectively bounded by the frame interval on the order of 0.1~s and the pixel resolution of the stabilized 8K imagery.

\subsubsection{Waymo Open Motion Dataset}

In WOMD, we compare the smoothed trajectories to the map geometry file. First, we extract the through-movement lanes upstream of the intersections and identify vehicles that are on those at the first data collection time step. Since the center line of each lane is represented by densely spaced Cartesian coordinates, we map the location of these vehicles onto the closest centerline point and sort them based on their distances to the end of the through-movement lane. Then, we use whether the most downstream vehicle is initially stopped as an indicator to determine the signal state and discard all through-movement vehicles where the first vehicle is initially moving. Finally, we iterate over the initially-stopped vehicles and obtain the time when their front bumper passes through the stop-line. The vehicles' front bumper locations can be obtained from the center locations, length, and direction of travel. If any of the following conditions are met, we discard that vehicle and upstream vehicles in the queue: (i) the vehicle never passes the stop line due to a data collection error, for example, some vehicles may not be consistently identified in the dataset, and (ii) the vehicle makes a turn after passing the stop line; this can be determined by extracting the next lane of the vehicle. 

Based on the criteria above, we have extracted a total of 1{,}152 lanes with discharging vehicles, where 180 lanes contain a Waymo vehicle.

\subsection{Discharge Headway Distribution}

\begin{figure}[htbp]
    \centering
    \begin{subfigure}[b]{0.45\textwidth}
        \includegraphics[width=\textwidth]{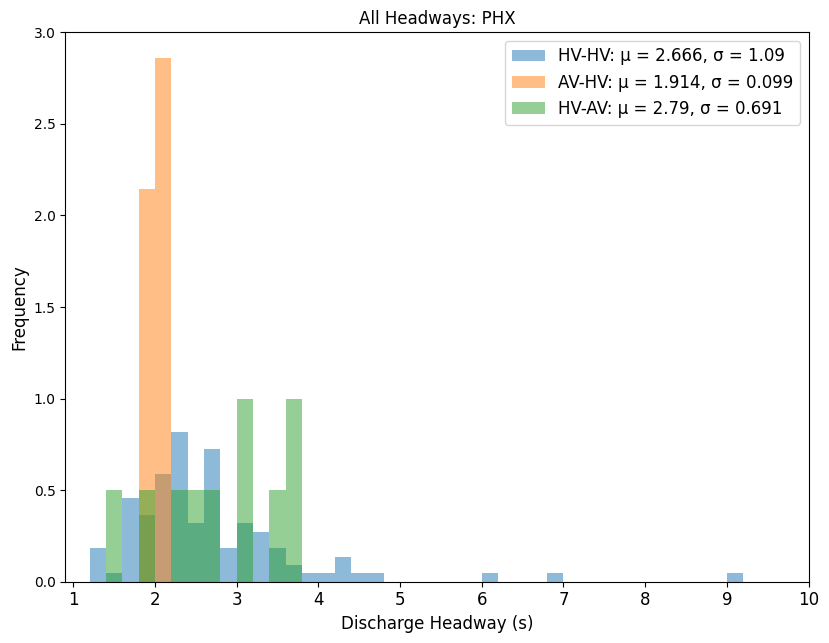}
        \caption{Phoenix Dataset}
        \label{hw_phx_overall}
    \end{subfigure}
    \hfill
    \begin{subfigure}[b]{0.45\textwidth}
        \includegraphics[width=\textwidth]{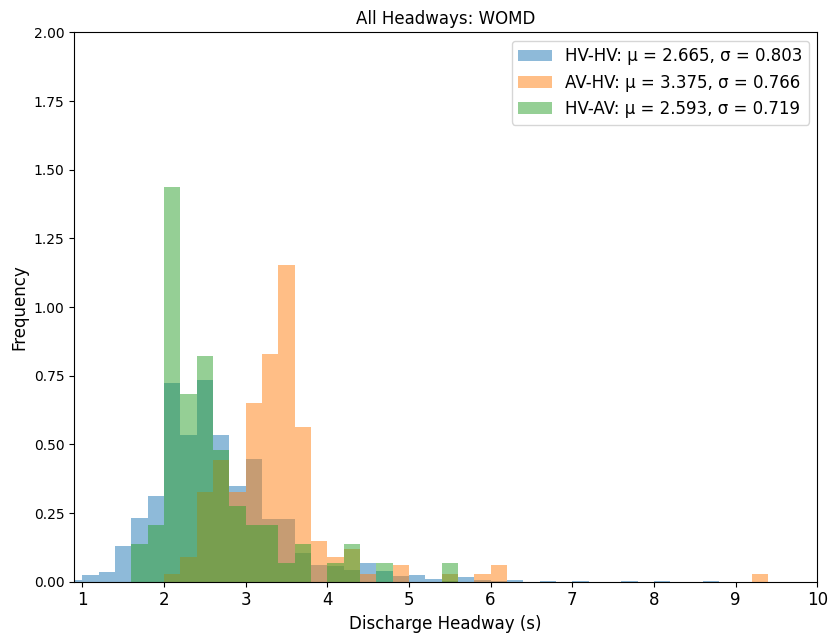}
        \caption{Waymo Open Dataset}
        \label{hw_}
    \end{subfigure}
    \caption{Overall Discharge Headway Distribution}
    \label{hw_overall}
\end{figure}

With the collected headway data, we compare their distribution across three cases: HV following HV (HV-HV), AV following HV (AV-HV), and HV following AV (AV-HV). The sample size across the two datasets differs: In the PHX dataset, the sample sizes of (HV-HV), (AV-HV), and (AV-HV) are 110, 7, and 10, respectively. In WOMD, these numbers are 1{,}788, 169, and 73, respectively. The overall headway distributions of the two datasets are shown in Figure \ref{hw_overall}. 

To compare these distributions statistically, we perform the Kolmogorov-Smirnov test \citep{massey1951kolmogorov}, which determines whether each case of discharge headway is drawn from the same headway distribution. The resulting p-values of HV-HV, AV-HV, and HV-AV cases are 0.125, 0.000, and 0.172, respectively, which means that under a 5\% confidence level, we can only conclude that the headway distributions of the AV-HV case are different between the datasets. Indeed, according to Fig.\ref{hw_overall}, AV-HV headway seems to be deterministic around 2s in the PHX dataset, which is lower than the lowest value observed in WOMD. This discrepancy indicates that WOMD might inaccurately capture AV trajectories or reflect outdated AV discharging behaviors, which are noticeably more conservative than those observed in the PHX dataset collected in the same year. 

% maybe state that WOMD is unreliable

\begin{figure}[!htb]
    \centering
    \includegraphics[width=0.7\textwidth]{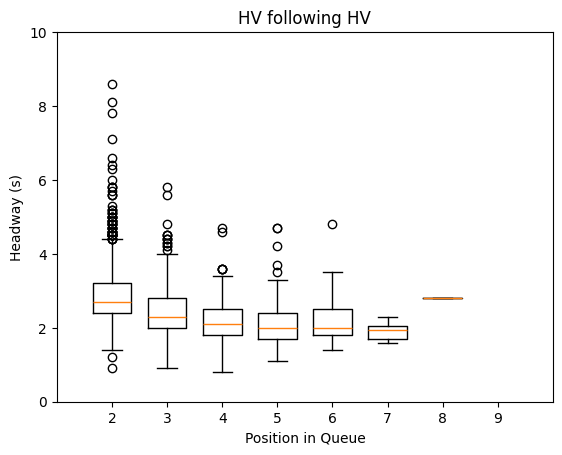} %
    \caption{Characteristics of queue discharge headways from WOMD}
    \label{hw_by_queue}
\end{figure}

While observing the overall headway distribution seems to suggest that the discharging behavior of HVs is similar between WOMD and PHX, examining the distribution of headway at different queue positions is important because the its mean tends to decrease for vehicles upstream of a queue \citep{lin2005headway, manual2000highway, greenshields1946traffic}. This phenomenon is called headway compression and is observed in WOMD for HV--HV case, as shown by  Fig.\ref{hw_by_queue}.

Due to the small PHX sample size, we compare the HV-HV headways in WOMD to those in Lin et al. \citep{lin2005headway}, where three intersections in the United States are included. Their reported results are reliable because they are obtained through traffic camera videos at a high resolution.

We conduct Welch's t-tests to determine whether the population means of the headways are equal for the two datasets at each queue position \citep{welch1947generalization}. Table \ref{headway-comparison} shows the sample headway distribution and testing results. We note that the headway statistics for Lin et al.\citep{lin2005headway} are aggregated across all intersections. The p-values are less than 0.05 for all positions, and the mean headways for WOMD are consistently higher. This indicates that WOMD may have overestimated the discharge headways for the HV-HV case.

\begin{table}[h!]
\centering
\begin{tabular}{||p{2.3cm}||p{1.5cm}|p{1.5cm}|p{1cm}||p{1.5cm}|p{1.5cm}|p{1cm}||p{1.2cm}|p{1.2cm}||} 
 \hline\hline
 \multirow{2}{*}{\textbf{Position}} 
 & \multicolumn{3}{c||}{\textbf{\cite{lin2005headway}}}  
 & \multicolumn{3}{c||}{\textbf{WOMD}} 
 & \multirow{2}{*}{\textbf{t-score}}
 & \multirow{2}{*}{\textbf{p-val}} \\

\cline{2-7}
 & \textbf{Mean} & \textbf{Std. Dev.} & \textbf{N} & \textbf{Mean} & \textbf{Std. Dev.} & \textbf{N} & & \\
 \hline\hline

% Example row (you can duplicate and edit these)
2 & 2.31 & 0.64 & 470 & 2.89 & 0.813 & 1031 & 14.72 & 0.000  \\

\hline

3 & 2.11 & 0.60 & 448 & 2.49 & 0.68 & 448 & 8.87 & 0.000  \\
\hline

4 & 1.99 & 0.65 & 430 & 2.15 & 0.57 & 207 & 3.202 & 0.001 \\
\hline

5 & 1.89 & 0.56 & 409 & 2.17 & 0.73 &  70 & 3.129 & 0.002 \\
\hline

6 & 1.86 & 0.55 & 393 & 2.21 & 0.72 & 23 & 2.261 & 0.033 \\
\hline
% ...

% I am done! Should I help read other parts?
 \hline\hline
\end{tabular}
\caption{Headway Comparison between WOMD and Lin et al.\citep{lin2005headway} by Position in Queue}
\label{headway-comparison}
\end{table}

% insert table

% overall distribution 

% This looks ugly and need to think about how to continue this

% 2, 3, 4+

% describe the distributions for each case
% Location dependent and therefore we compare the first 
% cite existing papers and state that our dataset has a value, especially for the HV to HV case.

We note that intersection discharge dynamics inherently capture the
acceleration regime of urban driving. The start-up process following a green
signal corresponds to the acceleration-from-stop phase, complementing the
deceleration-to-stop analysis presented later in Section~\ref{sec:decel}.

\section{Deceleration-to-Stop Behavior Analysis}
\label{sec:decel}
When comparing car-following behavior in PHX and WOMD, it is important to account for the high variability in human responses versus the relatively deterministic nature of AV's control. Therefore, in this section, we focus on the decelerating-to-stop phase of car-following, which captures salient control actions such as gap perception, signal anticipation, and braking. This phase offers a sensitive lens for validating whether WOMD reflects real-world AV behavior.

\subsection{Identification of Decelerating-to-Stop Cases}
To identify car-following episodes from both the PHX and WOMD datasets, we applied consistent extraction criteria to ensure a fair comparison. Specifically, we extracted car-following episodes with a minimum duration of 10 seconds, taking into account the 20-second maximum trajectory length available in the WOMD dataset. We limited the maximum distance between vehicles during the car-following to 50 meters and excluded the car-following episodes where the following vehicle's maximum speed during the episode did not exceed 3 $m/s$. This process yielded 143 car-following episodes from the PHX dataset and 47{,}826 pairs from the WOMD dataset. Among these, the WOMD scenarios contain 4{,}198 episodes where the following vehicle is flagged as the self-driving car (SDC) in WOMD metadata, whereas all ego vehicles in the PHX dataset correspond to the driverless ride-hail service described in Section~\ref{sec:phx_data}.

\begin{figure}[!htb]
    \centering
    \includegraphics[width=0.7\textwidth]{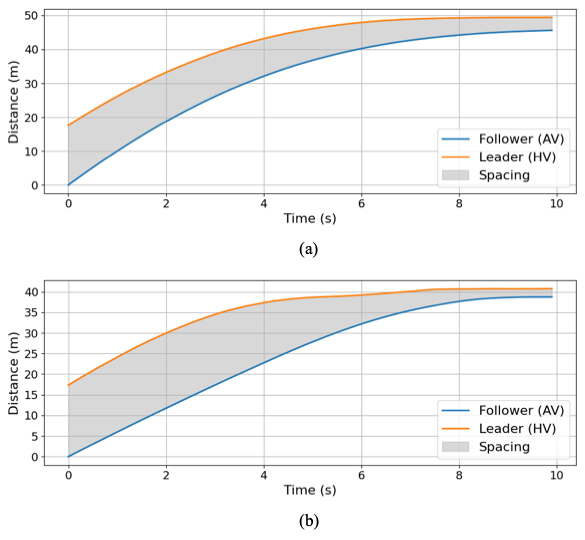}
    \caption{Example trajectories of extracted decelerating-to-stop segments from (a) PHX and (b)WOMD}
    \label{fig:trajectory_plot}
\end{figure}

Among the identified car following episodes, we focused on a specific subset in which the following vehicle decelerates to a stop behind the leader, extracting these segments using a specified process. The process begins by identifying time windows where both the follower and leader are moving slower than $\alpha$ for at least $\beta$ seconds. Once a stopping window for longer than $\beta$ seconds is found, we search backward in time to determine the onset of deceleration. We include 1 second from the onset of the stop of the following vehicle from this stopping window for each of the extracted segments of decelerating-to-stop cases. Each identified decelerating-to-stop segment includes up to $\gamma_{max}-1$ seconds of the follower vehicle's deceleration leading up to the stop, plus 1 second of the follower vehicle being stopped. The segments shorter than $\gamma_{min}$ are discarded. To confirm that the follower's stop results from the leading vehicle and not from external factors such as signals or pedestrians, segments in which the spacing exceeds $\delta$ during the stop are also discarded. We set the thresholds as $\alpha=1.0m/s$, $\beta=1.0s$, $\gamma_{min}=3.0s$, $\gamma_{max}=10.0s$, and $\delta=4.0m$. After applying the process, we extracted 38 valid decelerating-to-stop segments from the PHX dataset, comprising 3,434 time steps (each 0.1 second), and 775 segments from the WOMD, totaling 48,909 time steps. Since the same extraction process was applied for both datasets, we expect the decelerating-to-stop trajectories of Waymo vehicles to align across the datasets. Figures~\ref{fig:trajectory_plot}(a–b) show example trajectories of the extracted decelerating-to-stop segments from the PHX and WOMD datasets.

To avoid artifacts from occasional tracking failures in both datasets, we applied an outlier filter before the Markov and DTW analyses. Any car-following episode containing a time step with longitudinal acceleration magnitude exceeding $10~\mathrm{m/s^2}$ or an instantaneous change in spacing greater than $30~\mathrm{m}$ over a single $0.1$~s interval was discarded. This filter removed fewer than $1\%$ of candidate episodes in each dataset and did not significantly affect the summary statistics reported below.

\subsection{Car-following Comparison Between PHX and WOMD using Markov Decision Process}

To evaluate the behavior of Waymo vehicles in the PHX and WOMD datasets, we model the decelerating-to-stop dynamics as a Markov decision process. We treat the WOMD dataset as a high-fidelity behavioral reference, in the sense that it offers a large number of episodes collected with on-board sensing and proprietary tracking, but we do not assume that WOMD trajectories are entirely free of smoothing or perception error. In contrast, the PHX dataset exhibits empirically characterized error due to aerial videography and trajectory extraction, which we explicitly account for using the probabilistic methods described below. Given the extensive WOMD dataset, we train the transition matrix on WOMD and use it to assess whether PHX trajectories exhibit similar decelerating-to-stop behavior.

We first construct a transition matrix using the WOMD to represent the probability of transitioning between discretized bins during decelerating-to-stop segments. Each bin is defined by three variables, discretized as follows: $\Delta v$ ranges from -16 m/s to 16 m/s with 2 m/s intervals, $g$ from 0 to 32 m in 1 m intervals, and $v_f$ from 0 m/s to 16 m/s in 2 m/s intervals, resulting in a total of 4,096 bins.

For each decelerating-to-stop segment in the WOMD, we extract time-ordered sequences of bins based on the observed values of $\Delta v$, $g$, and $v_f$ at each 0.1-second time step. We count the number of transitions between consecutive bins across all segments to form the transition matrix. The transition matrix $P$ is then row-normalized so that each row sums to 1, where $P[m_t, m_{t+1}]$ represents the probability of transitioning from bin $m_t$ to bin $m_{t+1}$. To ensure the matrix is irreducible, we add a small value of $10^{-6}$ to all entries before normalizing.

To compare the PHX dataset against the WOMD, we evaluate the probability of each observed transition in the PHX dataset using the transition matrix $P$ trained on WOMD, incorporating the previously described normally distributed measurement errors present in the PHX data. For each transition in a PHX trajectory from time $t$ to $t+0.1$, we compute a probability distribution over possible states at both times based on the observed values and their error models, enabling a robust comparison to the WOMD.

% Based on the estimated error previously described, we model the relative speed error [m/s] as $\mathcal{N}(0, 0.0899^2)$; spacing error [m] as $\mathcal{N}(0, (\sqrt{2*0.0164^2} \cdot g)^2)$, corresponding to a standard deviation equal to 1.64\% spacing percentage error; and follower speed error [m/s] as $\mathcal{N}(0.0163, 0.0636^2)$. 

Specifically, for an observation at time $t$ with values $(\Delta v, g, v_{f})_t$, we compute a probability vector $q_t \in \mathbb{R}^{4096}$, assigning probabilities to each discretized bin defined over the three state variables. Each bin $i$ corresponds to a 3D interval, $[a_{\Delta v}, b_{\Delta v}) \times [a_g, b_g) \times [a_{v_f}, b_{v_f})$.

Assuming independence across the variables, the probability assigned to a given bin is computed as the product of the marginal probabilities from the corresponding Gaussian distributions. Formally, for each bin $i$, we define:

\begin{equation}
\label{eq:bin_probability}
q_t[i] =
\left[ \Phi_{\Delta v}(b_{\Delta v}) - \Phi_{\Delta v}(a_{\Delta v}) \right] \cdot
\left[ \Phi_g(b_g) - \Phi_g(a_g) \right] \cdot
\left[ \Phi_{v_f}(b_{v_f}) - \Phi_{v_f}(a_{v_f}) \right],
\end{equation}
\vspace{1mm}

\noindent where $\Phi_{\Delta v}$, $\Phi_g$, and $\Phi_{v_f}$ are the CDFs of the respective Gaussian error models for each variable. The above construction assumes that the three state variables are independent given the observation, so that the joint probability over a bin factorizes into the product of marginal probabilities. This independence approximation is reasonable in our setting because the empirically estimated correlations in the PHX error model are modest. A more general formulation would incorporate the full empirical covariance matrix when computing bin probabilities in $q_t$, which we leave as an extension for future work.

The same procedure is applied to compute $q_{t+1}$ from the observation at time $t + 0.1$. Given the WOMD transition matrix $P$, the probability of transitioning from time $t$ to $t+0.1$ is computed as:

\begin{equation}
\label{eq:transtion_probability}
P_t = \sum_{i=1}^{4096} \sum_{j=1}^{4096} q_t[i] \cdot P[i,j] \cdot q_{t+1}[j].
\end{equation}
\vspace{1mm}

\begin{figure}[!htb]
    \centering
    \includegraphics[width=0.7\textwidth]{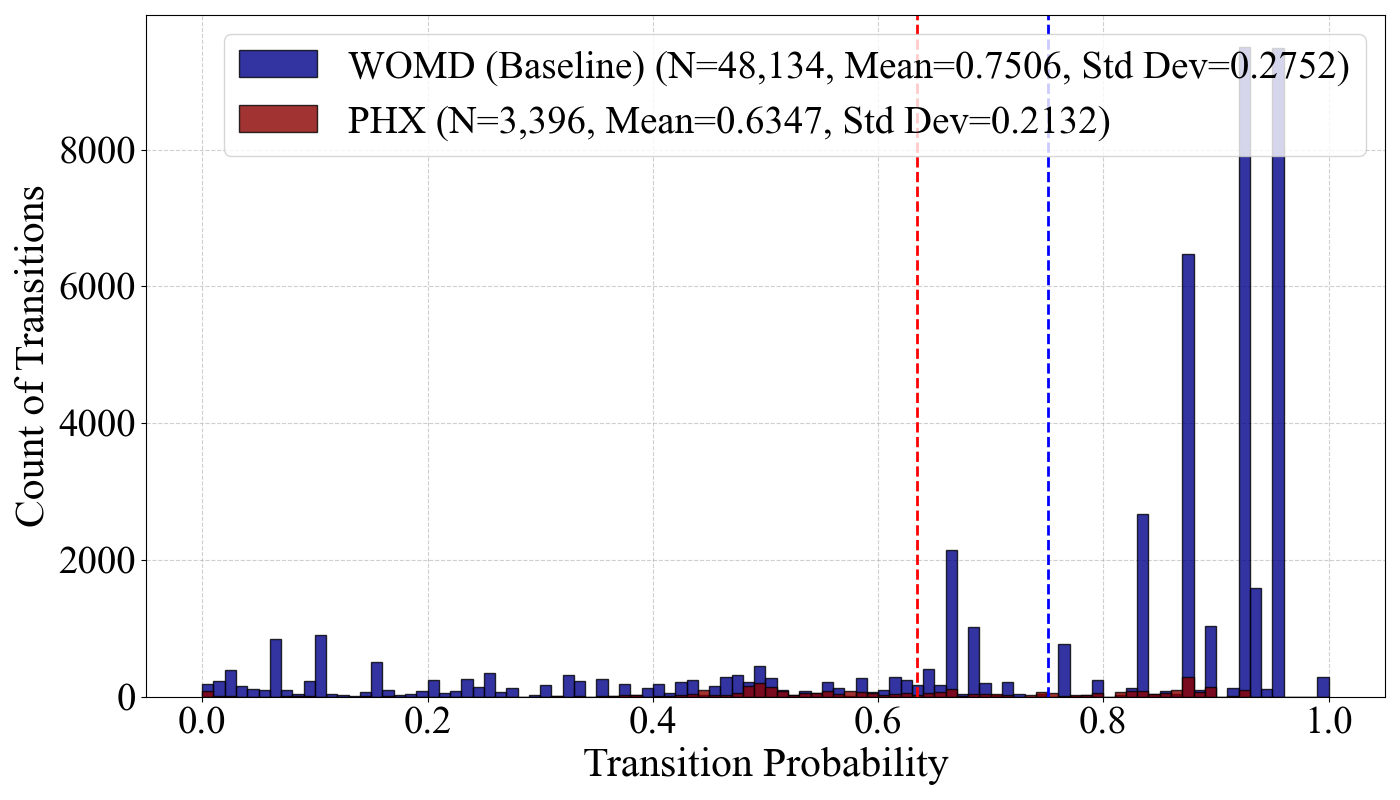}
    \caption{Transition probability histogram of PHX and WOMD segments using the WOMD transition matrix}
    \label{fig:prob_dist}
\end{figure}

Figure~\ref{fig:prob_dist} presents the distribution of individual transition probabilities across all decelerating-to-stop segments, evaluated using the WOMD-trained transition matrix. The PHX dataset shows a lower mean transition probability of 0.6347 compared to the WOMD baseline of 0.7506. The PHX distribution notably lacks the high-probability peaks observed in the WOMD dataset. In particular, WOMD exhibits prominent peaks above 0.6 and a dense concentration near 0.9, whereas the PHX distribution does not show a clear peak. However, both datasets include some low-probability transitions below 0.2, suggesting that the presence of noise or rare transitions is not unique to PHX. Therefore, based solely on this distribution, it is not conclusive whether the PHX decelerating-to-stop behavior substantially deviates from that of WOMD.

We further assess whether entire PHX decelerating-to-stop trajectories align with the car-following behavior modeled by the WOMD transition matrix by computing the geometric mean transition probability per step. Specifically, for a trajectory segment with $n$ time steps (i.e., $n - 1$ transitions), we compute:

\begin{equation}
\label{eq:geom_mean_combined}
\text{GeomMean} = \exp \left( \frac{1}{n - 1} \sum_{t=1}^{n - 1} \log P_t \right).
\end{equation}
\vspace{1mm}

This metric summarizes the average per-step transition probability under the WOMD model and provides a trajectory-level measure of how consistent each AV trajectory in the PHX dataset aligns with those in WOMD.

\begin{figure}[!htb]
    \centering
    \includegraphics[width=0.7\textwidth]{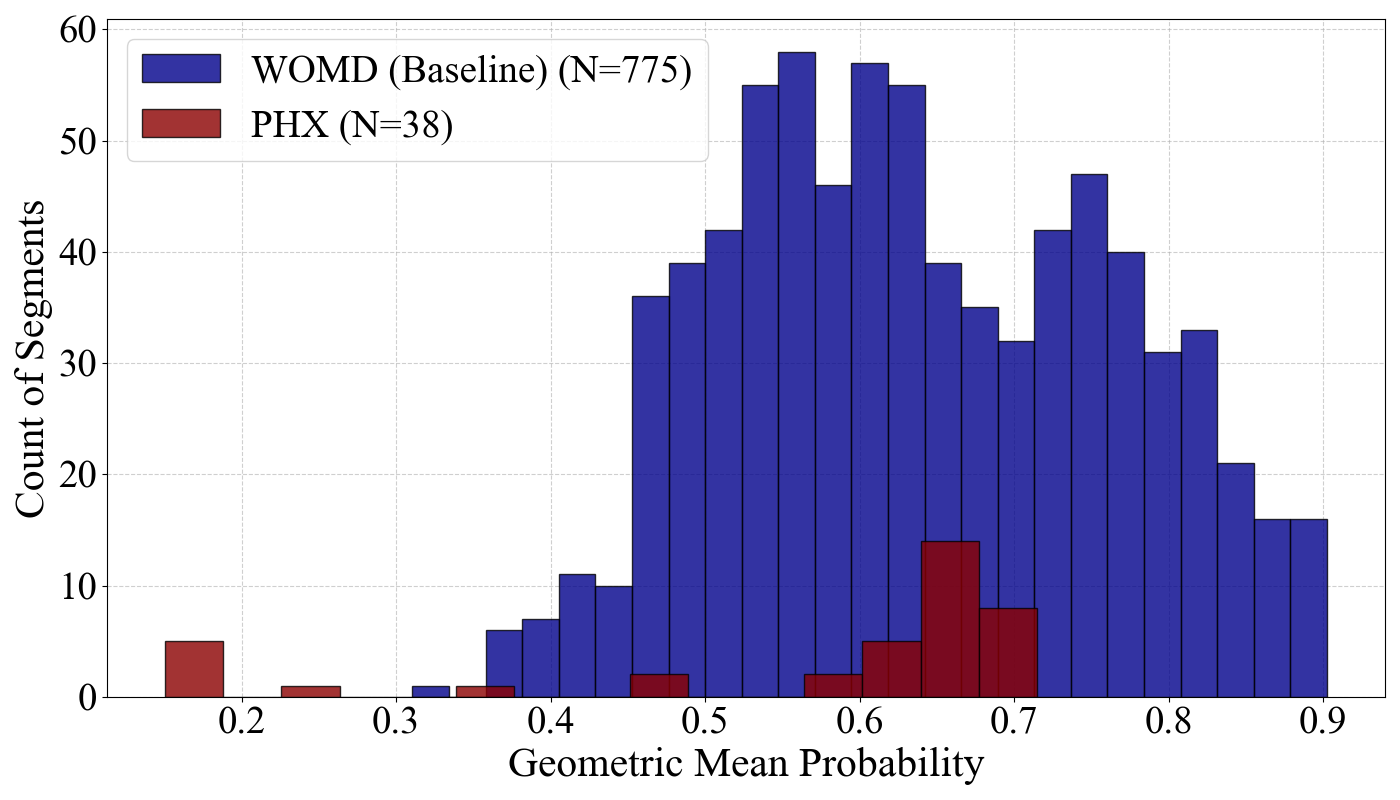}
    \caption{Geometric mean transition probabilities of PHX and WOMD trajectory segments}
    \label{fig:geom_prob_dist}
\end{figure}

Figure~\ref{fig:geom_prob_dist} further investigates the alignment of PHX behavior by evaluating the geometric mean transition probability per trajectory segment. This metric, again computed using the WOMD transition matrix, reveals a clearer separation between the datasets. While the WOMD trajectories show a concentrated and higher-probability distribution, 5 out of 38 PHX samples fall below a geometric mean of 0.3, indicating significant deviation from the behavior in WOMD. These low-probability segments cannot be plausibly explained by the WOMD-based transition dynamics. Given the sufficient sample size and well-formed distribution in WOMD, we conclude that some AV behaviors in the PHX dataset are different from those in the WOMD, even without the need for a formal statistical test.

\subsection{Simulation–Extrapolation Adjustment of Car-Following Dynamics Comparison Using PHX Error Distribution}

Trajectory data collected with onboard perception systems invariably contain measurement error. In this study, the PHX dataset comes with an empirically characterized error distribution, denoted $\widehat F_{\varepsilon}$, whereas WOMD is often used as if it were ground-truth data in the literature\citep{hu2022processing,hu2023IEEE}. In reality, WOMD trajectories are themselves the output of an on-board perception and temporal-smoothing pipeline, for which no explicit error characterization is available. This asymmetry implies that we can only apply SIMEX to the PHX dataset, for which an empirical error sample has been obtained, and must treat any latent measurement error in WOMD as part of the reference distribution against which we compare.

A formal bias correction procedure is required to determine whether the observed behavioral discrepancies between PHX and WOMD persist after the removal of PHX measurement noise. The SIMEX algorithm \citep{cook1994simulation} fulfills this role: it artificially inflates the known error, estimates the bias in a chosen statistic, and then extrapolates that bias back to zero error. If, at the error-free point, the PHX–WOMD distance remains substantially larger than WOMD's own internal variability, it suggests that the PHX dataset contains behavioral patterns absent from WOMD. 

SIMEX was introduced by \citet{cook1994simulation} to correct attenuation bias in linear‐regression coefficients measured with error. The method rapidly gained popularity in epidemiology, chemometrics, and econometrics, where instrument error is ubiquitous \citep{carroll1995measurement}. Empirical‐SIMEX, proposed by \citet{devanarayan2002empirical}, replaces the parametric error generator with a non‐parametric bootstrap from the empirical error sample, allowing heavy tails and skewness. Subsequent work has applied SIMEX to generalized linear models and survival analysis \citep{hardin2003simulation}, demonstrating its versatility wherever an explicit error distribution can be estimated.

For a statistic $T^{(0)}$ computed on noisy data, SIMEX proceeds in three stages: 

\textbf{(i) Simulation.} Generate pseudo datasets in which the measurement error variance is inflated by a multiplier $\lambda\ge 0$; obtain $T^{(\lambda)}$ for a grid $\Lambda\subset\mathbb R_{+}$. 

\textbf{(ii) Estimation.} Approximate $\mathbb E[T^{(\lambda)}]$ at each $\lambda$ by the Monte–Carlo mean over $B$ bootstrap replicates. 

\textbf{(iii) Extrapolation.} Fit a smooth function $\hat T(\lambda)$ through the points $\bigl\{(\lambda,\;T^{(\lambda)}) : \lambda\in\Lambda\bigr\}$ and evaluate it at $\lambda=-1$, where the total error variance $(1+\lambda)\operatorname{Var}(\varepsilon)$ vanishes.

Based on the above error distribution, the simulation step for SIMEX is performed as follows. Let $\lambda\in\Lambda=\{0,1,2\}$ and
$b=1,\dots,B$.
Define the pseudo-episode
\begin{equation}
\label{eq:sim_noise}
\tilde{\mathbf y}_{b}^{(\lambda)}(t)
  =\mathbf y^{\text{obs}}(t)
   +\sqrt{\lambda}\,\boldsymbol{\varepsilon}_{b}^{\ast}(t),
\quad
\boldsymbol{\varepsilon}_{b}^{\ast}(t)\sim\text{i.i.d. }\widehat F_{\varepsilon}.
\end{equation}

Then followed by the estimation step. Let
$T^{(\lambda)}
 =\tfrac{1}{B}\sum_{b=1}^{B}
     d_{\text{DTW}}\bigl(\tilde{\mathbf y}_{1,b}^{(\lambda)},
                         \tilde{\mathbf y}_{2,b}^{(\lambda)}\bigr)$.
Where $d_{DTW}$ is the distance based on Dynamic Time Warping. A quadratic fit is estimated
$\hat T(\lambda)=\beta_0+\beta_1\lambda+\beta_2\lambda^{2}$
to $\bigl\{\bigl(\lambda,T^{(\lambda)}\bigr):\lambda\in\Lambda\bigr\}$

Finally, the extrapolation step gives the bias‐free distance at $\lambda = -1$
\begin{equation}
\widehat d_{0}
 = \hat T(-1)
 = \beta_0-\beta_1+\beta_2.
\end{equation}

In order to compare the car-following behavior observed from PHX and WOMD, the car–following state vector is defined as

\begin{equation}
    \label{eq:cf_state_def}
\mathbf y(t)=\bigl[g(t),\;\Delta v(t),\;v(t)\bigr]^{\top},
\qquad
g=x_{\text{lead}}-x_{\text{ego}},\quad
\Delta v=v_{\text{lead}}-v_{\text{ego}}.
\end{equation}

Let the observed longitudinal position $x_i^{\text{obs}}(t)=x_i(t)+\varepsilon_{x,i}(t)$,
$i\in\{\text{ego},\text{lead}\}$, where
$\varepsilon_{x,i}(t)\sim\widehat F_{\varepsilon}$ are i.i.d.
Then we can first derive the corresponding error distribution for spacings, relative speeds, and speed as follows.
\begin{align}
g^{\text{obs}}(t)   &= g(t)+\bigl[\varepsilon_{x,\text{lead}}(t)-\varepsilon_{x,\text{ego}}(t)\bigr],\\
v^{\text{obs}}(t)   &= v(t)+{\varepsilon}_{v,\text{ego}}(t), \\
\Delta v^{\text{obs}}(t) &= \Delta v(t)
     +\bigl[{\varepsilon}_{v,\text{lead}}(t)-{\varepsilon}_{v,\text{ego}}(t)\bigr],
\end{align}

Consequently, the variance of spacing and relative speed can be derived as
    \label{eq:cf_state_cal}
    \begin{flalign}
            \operatorname{Var}\bigl[g^{\text{obs}}(t)\bigr]
       =2\,\sigma_{\varepsilon_x}^{2},\quad
\operatorname{Var}\bigl[\Delta v^{\text{obs}}(t)\bigr]
       =2\sigma_{\varepsilon_v}^{2}.
    \end{flalign}

Due to the fact that car-following cases vary in duration and driver reaction lag, a frame-wise Euclidean distance comparison would confound behavioral differences with temporal shifts in drivers' reactions. Dynamic Time Warping (DTW) resolves this by aligning two trajectories through a monotone warping path that minimizes cumulative local cost. While the classic formulation applies to scalar sequences \citep{sakoe1978dynamic,myers1980performance}, we extend it to the multivariate state vector \(\mathbf y(t)\) defined in Equation (\ref{eq:cf_state_def}).  Recent work shows that multivariate DTW captures driving behavior more robustly than single-variate measurements \citep{hosseini2022unsupervised,zhang2024lc}.

Let
\(
\mathbf A=[\mathbf a_{1},\dots,\mathbf a_{m}]
\) and
\(
\mathbf B=[\mathbf b_{1},\dots,\mathbf b_{n}]
\)
be two resampled episodes, where
\(\mathbf a_{i},\mathbf b_{j}\in\mathbb R^{3}\)
contain \(\bigl[g,\Delta v,v\bigr]^{\!\top}\) at frames \(i\) and \(j\).
Define the local cost matrix
\begin{equation}\label{eq:local_dist_matrix_multi}
      D\in\mathbb R^{m\times n},
      \qquad
      d_{ij}
      =\bigl\lVert \mathbf a_{i}-\mathbf b_{j}\bigr\rVert_{W}
      =\sqrt{\bigl(\mathbf a_{i}-\mathbf b_{j}\bigr)^{\!\top}W
                 \bigl(\mathbf a_{i}-\mathbf b_{j}\bigr)},
\end{equation}
where \(W=\operatorname{diag}\!\bigl\{\sigma_{g}^{-2},
                                       \sigma_{\Delta v}^{-2},
                                       \sigma_{v}^{-2}\bigr\}\)
standardizes the three channels by their pooled variance. The diagonal weighting matrix $W$ standardizes each channel by its pooled variance so that spacing, relative speed, and speed enter the local cost in comparable, dimensionless units. In principle, one could replace $W$ by the inverse of the full empirical covariance matrix, yielding a Mahalanobis-type distance that accounts for cross-channel correlations \citep{mahalanobis1936generalised}. We experimented with this alternative and found that it produces very similar rankings of DTW distances and, in particular, does not change the qualitative separation between PHX--WOMD and WOMD--WOMD pairs. Because the diagonal form is easier to interpret and less sensitive to covariance estimation error in small samples, we adopt it as the primary specification.

A warping path \(W=[(i_{1},j_{1}),\dots,(i_{K},j_{K})]\subset[1{:}m]\times[1{:}n]\)   must satisfy the boundary, continuity, and monotonicity conditions constraints of Sakoe–Chiba \citep{sakoe1978dynamic}.  
The multivariate DTW distance is therefore
\begin{equation}\label{eq:dtw_multi}
      DTW(\mathbf A,\mathbf B)
      =\min_{W}\;
        \Bigl\{\sum_{k=1}^{K} d_{i_{k}j_{k}}\Bigr\}^{1/2},
\end{equation}
which reduces to a shortest-path problem on the cumulative-cost matrix
and is solvable in \(O(mn)\) time \citep{senin2008dynamic}.  

Because the car-following episodes analyzed here have unequal length,
We report the length-normalized DTW score
\(
DTW^{\ast}=DTW/K,
\)
where \(K\) is the path length, following the normalization proposed by
\citep{zhang2024lc}.  
This normalized distance enters the SIMEX procedure described above; by construction, it is invariant to constant reaction-time offsets and robust to differing episode durations.

% It can be proven that this statistic is still unbiased. By construction,
% $\operatorname{Var}\bigl[
%   \sqrt{\lambda}\,\boldsymbol{\varepsilon}_{b}^{\ast}\bigr]
%   =(1+\lambda)\operatorname{Var}(\boldsymbol{\varepsilon})$.
% At $\lambda=-1$ the added variance is zero, hence
% $\tilde{\mathbf y}^{(-1)}\equiv\mathbf y^{\text{true}}$ almost surely.
% Under regularity conditions stated in
% Cook and Stefanski~\citep{cook1994simulation},
% $\mathbb E[T^{(\lambda)}]$ is
% analytic in $\lambda$ on $[-1,\infty)$,
% so $\hat T(-1)$ obtained from \eqref{eq:sim_noise}
% is a consistent estimator of
% $d_{\text{DTW*}}(\mathbf y_1^{\text{true}},\mathbf y_2^{\text{true}})$.
% \hfill$\square$

Next, this study computed for every PHX--WOMD pair to quantify cross‐dataset differences, and for every WOMD--WOMD pair to assess internal dataset variability. In total, we obtain $N_{\mathcal{PW}} = 38 \times 775 = 29,450$ cross-set distances and $N_{\mathcal{WW}} = \binom{775}{2} = 299,925$ internal distances. To statistically assess whether the PHX car-following cases lie outside the behavioral envelope of WOMD, we conduct a permutation test on the difference in mean SIMEX-corrected DTW distances.

Let $\mathcal{D}_\mathcal{PW} = \{d_1^{\mathcal{PW}},\dots,d_{n_1}^{\mathcal{PW}}\}$ and $\mathcal{D}_{\mathcal{WW}} = \{d_1^{\mathcal{WW}},\dots,d_{n_2}^{\mathcal{WW}}\}$ denote the sets of corrected distances between PHX--WOMD and WOMD--WOMD pairs, respectively, with $n_1 = 29,450$ and $n_2 = 299,925$. Define the observed test statistic as the difference in sample means:
\begin{equation}
    \bar{T}^{\text{CF}}_{\text{obs}} = \bar{d}^{CF}_{\mathcal{PW}} - \bar{d}^{CF}_{\mathcal{WW}},
\end{equation}
where $\bar{d}^{CF}_{\mathcal{PW}} = \frac{1}{n_1} \sum_{i=1}^{n_1} d_i^{\mathcal{PW}}$ and $\bar{d}^{CF}_{\mathcal{WW}} = \frac{1}{n_2} \sum_{j=1}^{n_2} d_j^{\mathcal{WW}}$. Under the null hypothesis that both sets are sampled from the same underlying behavioral distribution, we permute the labels of the pooled distances $\mathcal{D} = \mathcal{D}_{\mathcal{PW}} \cup \mathcal{D}_{\mathcal{WW}}$ and recompute the test statistic over $B = 5,000$ independent permutations. The $p$-value is computed as the empirical exceedance probability:
\begin{equation}
    p = \frac{1}{B} \sum_{b=1}^{B} \mathbb{I}\left\{T_b^\ast \geq T^{CF}_{\text{obs}}\right\},
\end{equation}
where $T_b^\ast$ denotes the permuted difference in means at iteration $b$, and $\mathbb{I}\{\cdot\}$ is the indicator function. This one-sided test corresponds to the hypothesis:
\begin{align}
    H_0 &:~ \mu^{CF}_{\mathcal{PW}} \leq \mu^{CF}_{\mathcal{WW}}, \\
    H_1 &:~ \mu^{CF}_{\mathcal{PW}} > \mu^{CF}_{\mathcal{WW}},
\end{align}

This directly tests whether the PHX–WOMD dissimilarities exceed those observed within WOMD itself.

\begin{figure}[!t]
  \centering
  \includegraphics[width=0.8\textwidth]{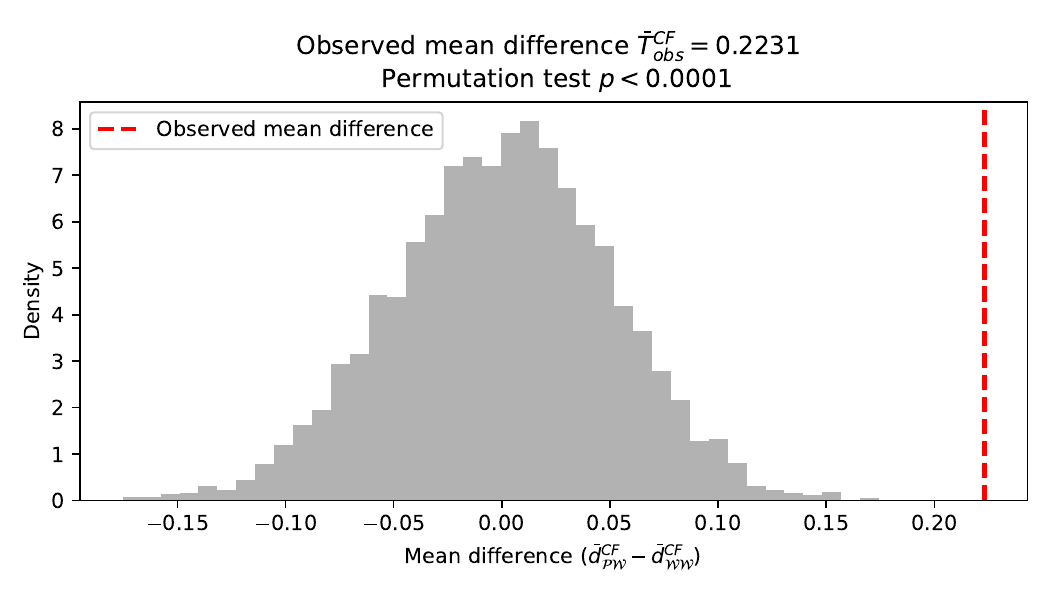}
  \caption{Permutation test distribution for the difference in SIMEX-corrected DTW distances between inter-dataset pairs (PHX--WOMD) and intra-dataset pairs (WOMD--WOMD) in car-following scenarios.}
  \label{fig:permutation_dtw_cf}
\end{figure}

The observed mean difference was $\bar{T}^{CF}_{\text{obs}} = 0.2231$. Across all $B = 5,000$ permutations, not a single iteration produced a difference in means as large as $T_{\text{obs}}$, yielding an empirical $p$-value of $p < 0.0001$. This strongly suggests that the PHX episodes are not simply sampled from the same behavioral regime as WOMD, but instead exhibit statistically significant deviations, even after adjusting for sensor noise in the PHX dataset. A histogram of the permutation distribution and the observed test statistic is shown in Figure~\ref{fig:permutation_dtw_cf}, further illustrating the magnitude of the deviation. These results reinforce the conclusion that the Waymo Open Motion Dataset does not fully capture the diversity of car-following behavior observed in naturalistic urban driving.

\section{Lane-changing Behavior Analysis} \label{sec:lanechange}

% Lane-changing maneuvers represent some of the most behaviorally complex and safety-critical actions in driving. Because of their interaction-rich nature—requiring coordination with lead and lag vehicles in both the current and target lanes, lane-changing behaviors are highly sensitive to the reliability of the trajectory \citep{zhang2024lc}. 

% The WOMD provides a large volume of lane-changing episodes based on perception, but the extent to which these trajectories reliably replicate naturalistic human behavior remains unclear. In this section, we compare lane-change trajectories extracted from WOMD to a naturalistic dataset collected independently in Phoenix. This section first details the extraction of lane-change events in the two datasets, then applies a SIMEX correction \citep{cook1994simulation} that incorporates the empirically estimated PHX error distribution. This two-step analysis enables a rigorous assessment of whether WOMD can closely mirror real-world lane-changing behavior once measurement error is properly accounted for.

Lane-changing is a complex, interaction-intensive maneuver that requires real-time coordination with surrounding vehicles. Analyzing such behavior using trajectory data demands high-quality input. Therefore, validating the quality of the trajectory data is essential before drawing behavioral conclusions. This study focuses on cases when AVs are performing the lane-changing maneuver, whose decision-making is generally less heterogeneous than that of human drivers. To ensure meaningful interaction, we further restrict our analyses to the cases where both lead and lag vehicles are present in the target lane. The absence of either vehicle may result from perception limitations rather than true gaps, and including such cases could confound interpretation.

\subsection{Identification of Lane-changing cases}

To identify lane-changing events from the WOMD, we employed a three-step procedure. First, we assigned each vehicle to a center-lane segment by identifying the closest center-lane point from the high-definition map. At each subsequent time step, the vehicle was reassigned to the nearest center lane among the following options: the current center lane, its exit connections, or its immediate left or right neighbors as defined in the map features.

Second, we filtered for lane-change events by identifying transitions where the assigned center-lane ID shifted to a left or right neighbor, excluding those directed to exit lanes. To remove maneuvers occurring within intersections, we excluded cases where the assigned center lane was an interpolated segment, indicating a virtual path through an intersection. We also filtered out lane changes that immediately followed right or left turns, where heading variation could confound classification. Only lane changes with a heading angle change of less than 0.2 radians before and after the maneuver were retained. This filtering process enabled consistent, automated identification of midblock lane-change events across all scenarios.

\begin{figure}[!htb]
    \centering
    \includegraphics[width=0.7\textwidth]{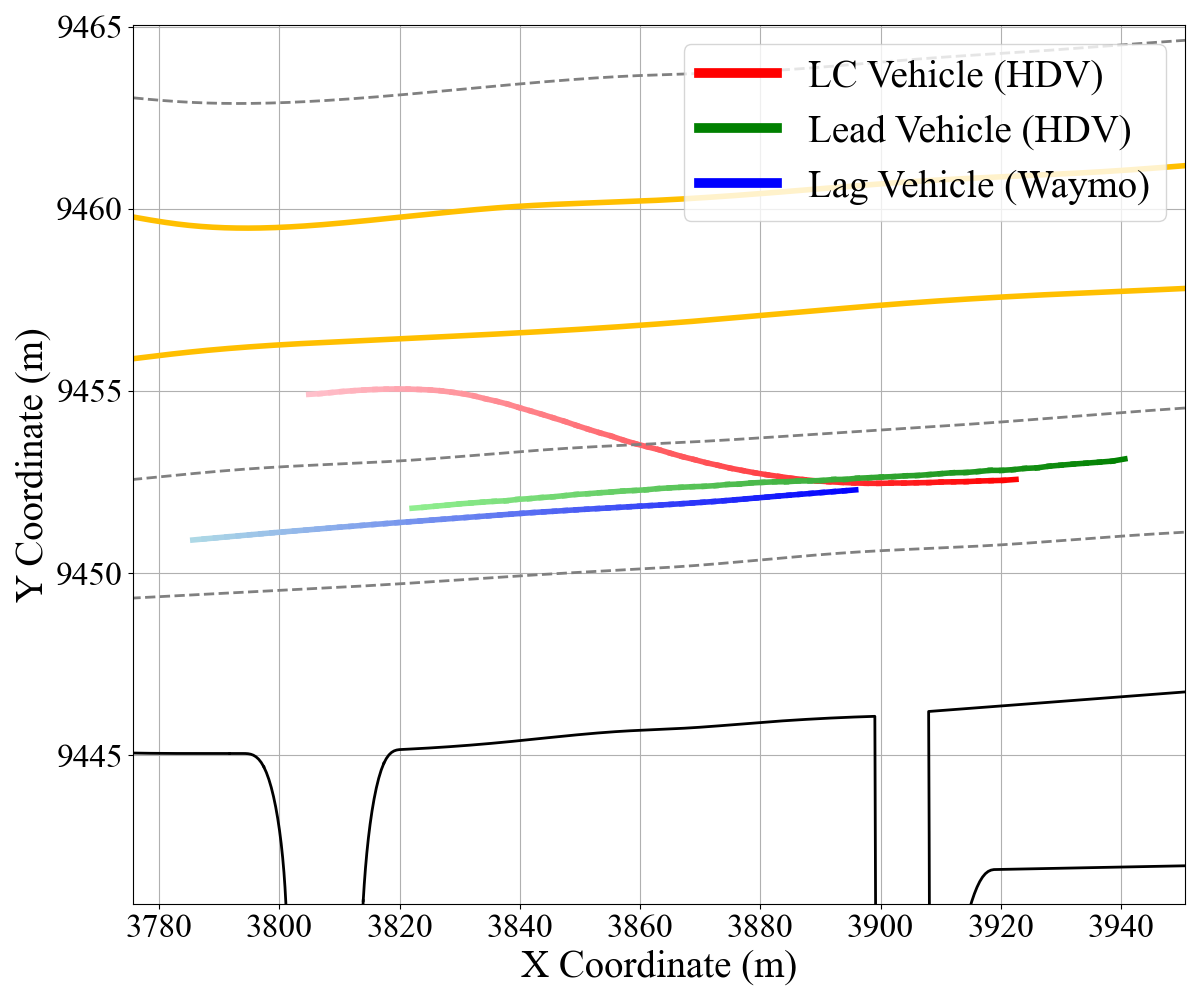}
    \caption{Example of a lane-changing trajectory from WOMD}
    \label{trajectory_plot}
\end{figure}

Third, we assigned lead and lag vehicles to all vehicles at each time step to support the gap acceptance analysis described in the next section. In urban scenarios, where center lanes often split into multiple successors, for example, at a shared right-turn/straight lane, multiple leader candidates may exist. We selected the leader with the shortest longitudinal distance along the assigned center line. Figure~\ref{trajectory_plot} shows an example trajectory of a lane-changing vehicle along with its lead and lag vehicles. The plot shows a $6$-second window centered at the frame when the lane-changing vehicle crosses the lane marking ($t=0$). Longitudinal position is measured along the lane centerline, and lateral position represents cross-lane displacement relative to the initial lane. 

For the PHX dataset, lane assignment was performed by manually segmenting the reference image into lane regions based on visible pavement markings. Lane changes were identified by detecting when the vehicle centroid crossed a lane boundary. For each lane-change event, we extracted a six-second window centered on the crossing, plotted the subject and surrounding trajectories, and calculated lead and lag gaps along with relative speed profiles. Because the error distribution in the PHX dataset is empirically estimated, it serves as a naturalistic benchmark for evaluating the reliability of WOMD lane-change trajectories in subsequent analyses.

We applied the same outlier filter as in the car-following analysis: lane changes whose six-second window contained longitudinal accelerations larger than $10~\mathrm{m/s^2}$ in magnitude or spacing jumps greater than $30~\mathrm{m}$ over $0.1$~s were excluded. The final sample consists of 13 PHX lane-change episodes and 200 comparator episodes from WOMD. In the PHX dataset, the ego vehicle in all analyzed lane-change events is a commercial robotaxi, while all surrounding vehicles are human-driven. During our observation period, we did not encounter any lane changes in which the lead or lag vehicle in the target lane was also an automated vehicle. Consequently, our lane-changing analysis reflects mixed traffic with a single AV ego rather than AV--AV interactions, which remain rare in current deployments.

\subsection{Simulation–Extrapolation Adjustment of Lane–Changing Dynamics Comparison Using the PHX Error Distribution}
\label{subsec:simex_lc}

Lane changes combine pronounced lateral motion with longitudinal interaction and are therefore an ideal testbed for evaluating the behavioural fidelity of WOMD after noise has been removed from the PHX reference data. We apply exactly the empirical–SIMEX procedure and multivariate DTW distance defined for car–following, but with a six–component lane–change state vector and the corresponding propagation of longitudinal and lateral error.

For every lane–change episode, we extract a \([-3\,\text{s},+3\,\text{s}]\) window centered on the frame when the centroid of the ego vehicle crosses the lane marking and represent the lane-changing case as
\begin{equation}
\mathbf s(t)=
\bigl[
    \Delta x(t),
    \Delta y(t),
    g_{L}(t),
    g_{F}(t),
    \Delta v_{L}(t),
    \Delta v_{F}(t)
\bigr]^{\!\top},
\end{equation}
where \(\Delta x\) and \(\Delta y\) are the ego-vehicle displacements relative to the initiation point, \(g_{L}=x_{\text{lead}}-x_{\text{ego}}\) and \(g_{F}=x_{\text{ego}}-x_{\text{lag}}\) are the lead and lag spacings, and \(\Delta v_{L}=v_{\text{lead}}-v_{\text{ego}}\) as well as \(\Delta v_{F}=v_{\text{ego}}-v_{\text{lag}}\) are the corresponding relative speeds.  Longitudinal and lateral positions inherit additive error \(x_i^{\mathrm{obs}}(t)=x_i(t)+\varepsilon_{x,i}(t)\) and \(y_i^{\mathrm{obs}}(t)=y_i(t)+\varepsilon_{y,i}(t)\) with \(\varepsilon_{x,i},\varepsilon_{y,i}\stackrel{\text{i.i.d.}}{\sim}\widehat F_{\varepsilon}\), respectively.  Consequently \(g_{L}^{\mathrm{obs}}=g_{L}+(\varepsilon_{x,\text{lead}}-\varepsilon_{x,\text{ego}})\) and \(g_{F}^{\mathrm{obs}}=g_{F}+(\varepsilon_{x,\text{ego}}-\varepsilon_{x,\text{lag}})\), so that each spacing channel acquires an additive error of variance \(2\sigma_{\varepsilon_x}^{2}\).  The two relative-speed channels inherit the first-difference error term already derived for car–following, with variance scaling \(2\sigma_{\varepsilon_v}^{2}\).

Pseudo episodes for SIMEX are generated by adding \(\sqrt{\lambda}\,\boldsymbol\varepsilon^{\ast}(t)\) to every position channel, where \(\lambda\in\{0,1,2\}\) and \(\boldsymbol\varepsilon^{\ast}\) is a bootstrap draw from \(\widehat F_{\varepsilon}\).  Each pseudo trajectory is converted back to the six-channel form \(\mathbf s(t)\). Similar to the car-following analysis, a multivariate DTW with a Sakoe–Chiba band of \(\pm 2\) frames computes the length-normalised distance \(DTW^{\ast}\).  The bootstrap means \(T^{(\lambda)}\) are fitted with the quadratic \(\hat T(\lambda)=\beta_0+\beta_1\lambda+\beta_2\lambda^{2}\), and the bias-free distance is obtained at the zero-variance limit as \(\widehat d_{0}=\hat T(-1)=\beta_0-\beta_1+\beta_2\).

The SIMEX-corrected DTW distance $\widehat{d}_{0}$ is then computed for every PHX--WOMD episode pair to quantify cross-dataset divergence in lane-changing behavior, and for every WOMD--WOMD pair to establish an internal benchmark. In total, we obtain $N_{\mathcal{PW}} = 13 \times 200 = 2{,}600$ inter-dataset distances and $N_{\mathcal{WW}} = \binom{200}{2} = 19{,}900$ intra-dataset distances. A one-sided permutation test is employed to determine whether the cross-set distances are significantly larger than the internal variability of WOMD.

Let $\mathcal{D}_{\mathcal{PW}} = \{d_1^{\mathcal{PW}}, \dots, d_{n_1}^{\mathcal{PW}}\}$ and $\mathcal{D}_{\mathcal{WW}} = \{d_1^{\mathcal{WW}}, \dots, d_{n_2}^{\mathcal{WW}}\}$ be the SIMEX-corrected DTW distances for PHX--WOMD and WOMD--WOMD episode pairs, respectively, with $n_1 = 2{,}600$ and $n_2 = 19{,}900$. The observed test statistic is the difference in mean DTW distances:
\begin{equation}
    \bar{T}^{\text{LC}}_{\text{obs}} = \bar{d}^{\text{LC}}_{\mathcal{PW}} - \bar{d}^{\text{LC}}_{\mathcal{WW}},
\end{equation}
where the sample means are given by $\bar{d}^{\text{LC}}_{\mathcal{PW}} = \frac{1}{n_1} \sum_{i=1}^{n_1} d_i^{\mathcal{PW}}$ and $\bar{d}^{\text{LC}}_{\mathcal{WW}} = \frac{1}{n_2} \sum_{j=1}^{n_2} d_j^{\mathcal{WW}}$. Under the null hypothesis that the PHX and WOMD lane-change behaviors arise from the same underlying distribution, the labels of the pooled set $\mathcal{D} = \mathcal{D}_{\mathcal{PW}} \cup \mathcal{D}_{\mathcal{WW}}$ are randomly permuted $B = 5{,}000$ times. For each permutation, we recompute the difference in means, yielding a null distribution $\{T_1^\ast, T_2^\ast, \dots, T_B^\ast\}$. The empirical $p$-value is then calculated as
\begin{equation}
    p = \frac{1}{B} \sum_{b=1}^{B} \mathbb{I}\left\{T_b^\ast \geq \bar{T}^{\text{LC}}_{\text{obs}} \right\},
\end{equation}
corresponding to the one-sided hypothesis test
\begin{align}
    H_0 &:~ \mu^{\text{LC}}_{\mathcal{PW}} \leq \mu^{\text{LC}}_{\mathcal{WW}}, \\
    H_1 &:~ \mu^{\text{LC}}_{\mathcal{PW}} > \mu^{\text{LC}}_{\mathcal{WW}}.
\end{align}

\begin{figure}[!t]
  \centering 
  \includegraphics[width=0.8\textwidth]{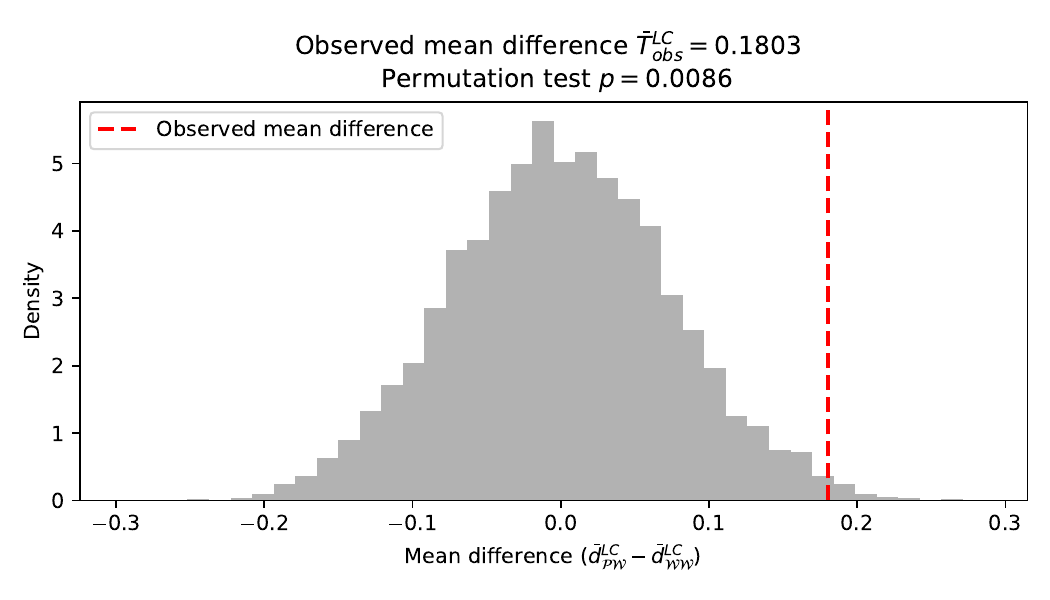}
  \caption{Permutation test distribution for the difference in SIMEX-corrected DTW distances between inter-dataset lane-changing pairs (PHX--WOMD) and intra-dataset pairs (WOMD--WOMD).}
  \label{fig:permutation_dtw_lc}
\end{figure}

The observed mean difference was found to be $\bar{T}^{\text{LC}}_{\text{obs}} = 0.1803$. The permutation test yielded a $p$-value of $p  = 0.0086$, indicating that the cross-set distances are statistically larger than the internal variability of WOMD. %Given that this conclusion is based on only 13 PHX lane changes matched against 200 WOMD episodes, the result should be interpreted as evidence that PHX behaviors occupy the upper tail of WOMD's internal distance distribution, rather than as a precise estimate of the magnitude of the difference. 
The main message is that several everyday lane changes observed in PHX are more dissimilar to any WOMD lane change than typical WOMD lane changes are to one another. We emphasize that this conclusion is based on 13 lane-changing episodes from PHX and 200 comparator episodes from WOMD; while the permutation test p-value indicates a statistically detectable difference at this sample size, the effect size and qualitative shape of the distance distributions are more informative than the exact p-value. Figure~\ref{fig:permutation_dtw_lc} visualizes the null distribution of mean differences and the observed test statistic. These results suggest that the lane-changing dynamics observed in PHX do not lie within the behavioral envelope of WOMD, despite accounting for distance and velocity error in the PHX measurements. Thus, caution is warranted when using WOMD to model lane-changing behavior.

\section{Conclusion} \label{sec:conclusion}
\textcolor{black}{This study introduced a robust methodology for validating trajectory datasets using an independent naturalistic ground-truth dataset with quantified error distribution. By combining manual, frame-anchored ground truth for intersection discharge headways and SIMEX-based correction and distributional tests for car-following and lane-changing behaviors, the framework enables validity claims that do not depend on any post-processing of the public dataset. This independent validation perspective complements earlier trajectory quality studies that assessed datasets such as NGSIM primarily through internal diagnostics and filtered-trajectory references.} %\citep{punzo2011assessment,montanino2013making,montanino2015trajectory}.}

Particularly, this study critically examined the extent to which WOMD can be used in driving behavior analysis by comparing its trajectory data to an independently collected dataset of commercial robotaxi operations in Phoenix, AZ (PHX). We focused on three scenarios: intersection discharging, decelerating-to-stop, and lane-changing. Because the two datasets differ in collection period, operational design domain, and sensing modality, any discrepancies we observe necessarily conflate differences in control policy, roadway environment, and post-processing pipelines. Our analysis, therefore, does not aim to isolate a single causal factor, but instead asks a more practical question: do the behaviors observed in recent, naturalistic Level~4 operations fall inside the behavioral envelope spanned by WOMD trajectories?

Across all scenarios, we found a significant difference between the behaviors observed in WOMD and those captured in PHX. WOMD consistently underrepresents short headways and sharp decelerations. Because the PHX discharge headways are anchored in frame-by-frame manual annotations of vehicles crossing the stop line, with uncertainty at the level of a single video frame, these discrepancies cannot be explained away as artefacts of aerial measurement error; they instead point to genuine differences between the naturalistic acceleration behavior we observe in recent robotaxi operations and the start-up dynamics represented in WOMD. These differences exist even after correcting for measurement error in PHX, suggesting that the differences stem not from data quality but from underlying behavioral characteristics. Permutation tests confirm that these cross-dataset differences are statistically significant and lie outside the internal variability of WOMD. 

These findings challenge the common assumption that WOMD provides ground-truth behavioral data suitable for calibration or validation of microscopic driving models. We caution researchers against using WOMD as an unquestioned benchmark for behavioral analysis, especially in safety-critical or interaction-rich contexts. Future research should incorporate independent validation datasets and consider smoothing artifacts when modeling driving behavior from open-source datasets derived from onboard sensor perceptions, such as WOMD.

More broadly, our study illustrates both the value and the limitations of cross-source validation for AV trajectory datasets. Because PHX and WOMD differ in operational design domain, geography, time period, and sensing stack, we can only conclude that certain patterns observed in recent naturalistic operations, such as short headways and sharp decelerations, are underrepresented in the WOMD dataset. We cannot uniquely attribute the discrepancy to a specific layer of the AV perception stack or to a particular post-processing choice. Future validation efforts will benefit from coordinated data releases that include both raw sensor streams and independent external measurements in overlapping time and space.

\section{Acknowledgements}
This work was supported by the National Science Foundation under Grant No. 2047937.

% \section{Acknowledgments}
\section{Declaration of generative AI and AI-assisted technologies in the writing process}
The authors acknowledge the use of AI-assisted tools (such as ChatGPT) for language editing and grammar refinement during manuscript preparation. No AI tool was used for generating novel content, data analysis, or drawing conclusions. All responsibility for the accuracy and integrity of the manuscript remains with the authors.

\section{AUTHOR CONTRIBUTIONS}
The authors confirm contribution to the paper as follows: study conception and design: Y. Zhang, S. Chung, N. Li, D. Monzer, A. Talebpour, H. S. Mahmassani and S. Hamdar; data collection: Y. Zhang, A. Talebpour, H. S. Mahmassani and S. Hamdar; analysis and interpretation of results: Y. Zhang, S. Chung, N. Li, D. Monzer, A. Talebpour, H. S. Mahmassani, and S. Hamdar; draft manuscript preparation: Y. Zhang, S. Chung, N. Li, D. Monzer, A. Talebpour, H. S. Mahmassani, and S. Hamdar. All authors reviewed the results and approved the final version of the manuscript.

\section{DECLARATION OF CONFLICTING INTERESTS}
The authors declared no potential conflicts of interest with respect to the research, authorship, and/or publication of this article.

\bibliographystyle{apalike}
\bibliography{trb_template}

\end{document}